\documentclass[runningheads]{llncs}

\usepackage[T1]{fontenc}
\usepackage{graphicx,verbatim}
\usepackage{url}
\usepackage{multirow}
\usepackage{amsmath}
\usepackage{makecell}
\usepackage{tabularx}

\begin{document}

% \title{Scaling Diverse Video Pretraining for Surgical Understanding}
\title{Scaling Video Pretraining for Surgical \\ Foundation Models}

\titlerunning{Scaling Diverse Medical Video Pretraining}

\author{
Sicheng Lu\inst{1}\textsuperscript{*} \and
Zikai Xiao\inst{2}\textsuperscript{*} \and
Jianhui Wei\inst{3}\textsuperscript{*} \and
Danyu Sun\inst{3} \and
Qi Lu\inst{4} \and
Keli Hu\inst{5} \and
Yang Feng\inst{6} \and
Jian Wu\inst{2} \and
Zongxin Yang\inst{7} \and
Zuozhu Liu\inst{3}
}
\authorrunning{Lu et al.}
\institute{
The Johns Hopkins University, United States
\and Zhejiang University, China
\and Zhejiang University-University of Illinois Urbana-Champaign Institute, China
\and Zhejiang Lab, China
\and Shaoxing University, China
\and Angelalign, China
\and Harvard Medical School, United States\\[2mm]
\textsuperscript{*} Equal contribution.
}

\maketitle

\begin{abstract}
Surgical video understanding is essential for computer-assisted interventions, yet existing surgical foundation models remain constrained by limited data scale, procedural diversity, and inconsistent evaluation, often lacking a reproducible training pipeline. We propose \textbf{SurgRec}, a scalable and reproducible pretraining recipe for surgical video understanding, instantiated with two variants: \textbf{SurgRec-MAE} and \textbf{SurgRec-JEPA}. 
We curate a large multi-source corpus of 10,535 videos and 214.5M frames spanning endoscopy, laparoscopy, cataract, and robotic surgery. 
Building on this corpus, we develop a unified pretraining pipeline with balanced sampling and standardize a reproducible benchmark across 16 downstream datasets and four clinical domains with consistent data splits. Across extensive comparisons against SSL baselines and vision-language models, SurgRec consistently achieves superior performance across downstream datasets. In contrast, VLMs prove unreliable for fine-grained temporal recognition, exhibiting both performance gaps and sensitivity to prompt phrasing. Our work provides a reproducible, scalable foundation for the community to build more general surgical video models. All code, models, and data will be publicly released.

\end{abstract}

\section{Introduction}\label{sec:intro}
Surgical video understanding is central to computer-assisted interventions, supporting a wide range of tasks from workflow recognition and skill assessment to intra-operative decision support. Although public benchmarks have accelerated progress~\cite{endonet,m2cai_lapchole,jigsaws,cataract101,kvasir_capsule}, supervised approaches remain constrained by costly temporal annotations and limited generalization across sites and procedures.

Self-supervised learning (SSL) offers a scalable path to transferable video representations without costly annotations. In the general domain, masked reconstruction (e.g., VideoMAE~\cite{videomae}), predictive latent modeling (e.g., V-JEPA~\cite{vjepa2}), and large-scale image SSL (e.g., DINO~\cite{dinov3}) have proven highly effective. Building on this momentum, surgical foundation models such as Endo-FM, SurgVLP, SurgeNetXL, SurgVISTA, and UniSurg~\cite{endofm,surgvlp,surgenetxl,surgvista,unisurg} have emerged, suggesting that broader procedure coverage and larger corpora can improve cross-domain transferability. Yet existing models are often limited in data scale and procedural diversity, and evaluations remain inconsistent across works, making it hard to draw reliable conclusions~\cite{surgbench}. Vision-language models (VLMs)~\cite{clip,llava15,llava_next_repo,qwen25vl,qwen3vl} present another appealing direction, but their reliability on fine-grained surgical recognition remains unclear, with performance often sensitive to prompt phrasing. These limitations highlight the need for a scalable pretraining recipe that can be rigorously and reproducibly evaluated across heterogeneous surgical domains.

In this work, we introduce \textbf{SurgRec}, a scalable and reproducible pretraining recipe for general surgical video understanding. We curate a heterogeneous corpus spanning endoscopy, laparoscopy, cataract, robotic surgery, and mixed surgical videos, totaling 10,535 videos and 214.5M frames: 2,790 videos (39.9M frames) from 32 public datasets and 7,745 videos (174.6M frames) from publicly accessible resources. Under a unified pipeline (Sec.~\ref{sec:pretrain_impl}), we instantiate \textbf{SurgRec-MAE} and \textbf{SurgRec-JEPA}, two surgical foundation model variants based on masked reconstruction and latent prediction, and evaluate them on a standardized multi-procedure benchmark with consistent data splits (Sec.~\ref{sec:benchmark_suite}), with VLM baselines included for comparison. Through this evaluation, we demonstrate that large-scale in-domain pretraining yields consistently strong performance across a wide range of downstream benchmarks, reflecting good generalization over diverse surgical procedures. Furthermore, our results reveal that VLMs struggle with both performance and 
stability in fine-grained surgical understanding.

Our main contributions are three-fold: (1) we curate a large, multi-source surgical video corpus of 10,535 videos and 214.5M frames spanning four major clinical domains, ensuring broad and long-tail procedural coverage; (2) we present SurgRec-MAE and SurgRec-JEPA, two surgical foundation model variants trained under a unified and reproducible pipeline; and (3) we establish a standardized multi-procedure benchmark with consistent data splits, evaluating both domain-pretrained encoders and VLM baselines on identical test sets.

\section{Large-Scale Pretraining Corpus}
\subsection{Data Sources and Composition}
Our pretraining corpus spans diverse surgical procedures, imaging modalities, and clinical sites, with full temporal context preserved across videos. We source data from three streams: (i) public clinical datasets, (ii) web-crawled surgical videos, and (iii) private data to improve coverage of procedure types.

\begin{figure}[t]
\centering
\includegraphics[width=0.8\linewidth]{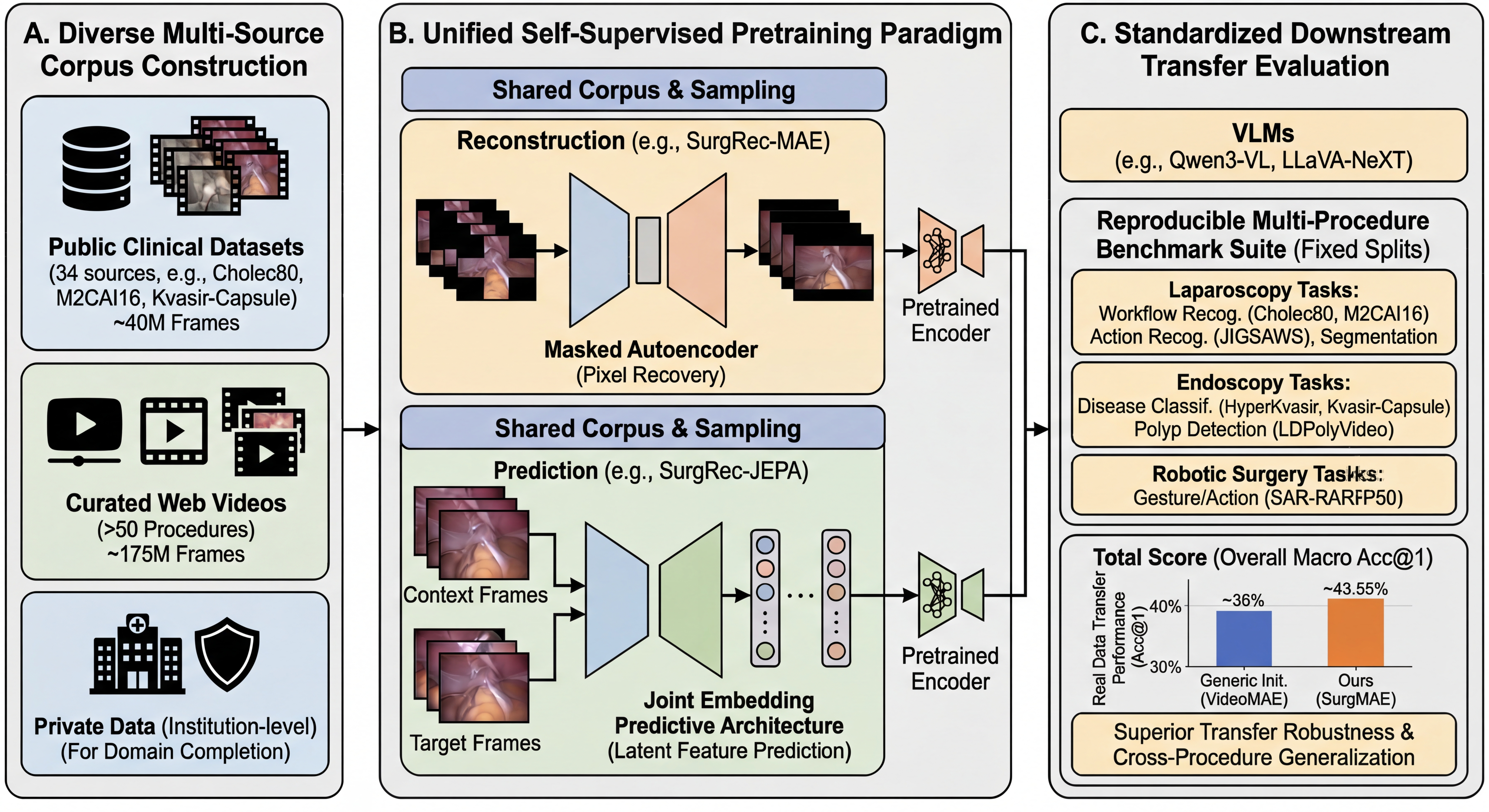}
\caption{Overview of the proposed medical video pretraining and evaluation framework.}
\label{fig:corpus_overview}
\end{figure}

\begin{figure}[t]
    \centering
    \begin{minipage}[b]{0.42\textwidth}
        \centering
        \includegraphics[width=\textwidth]{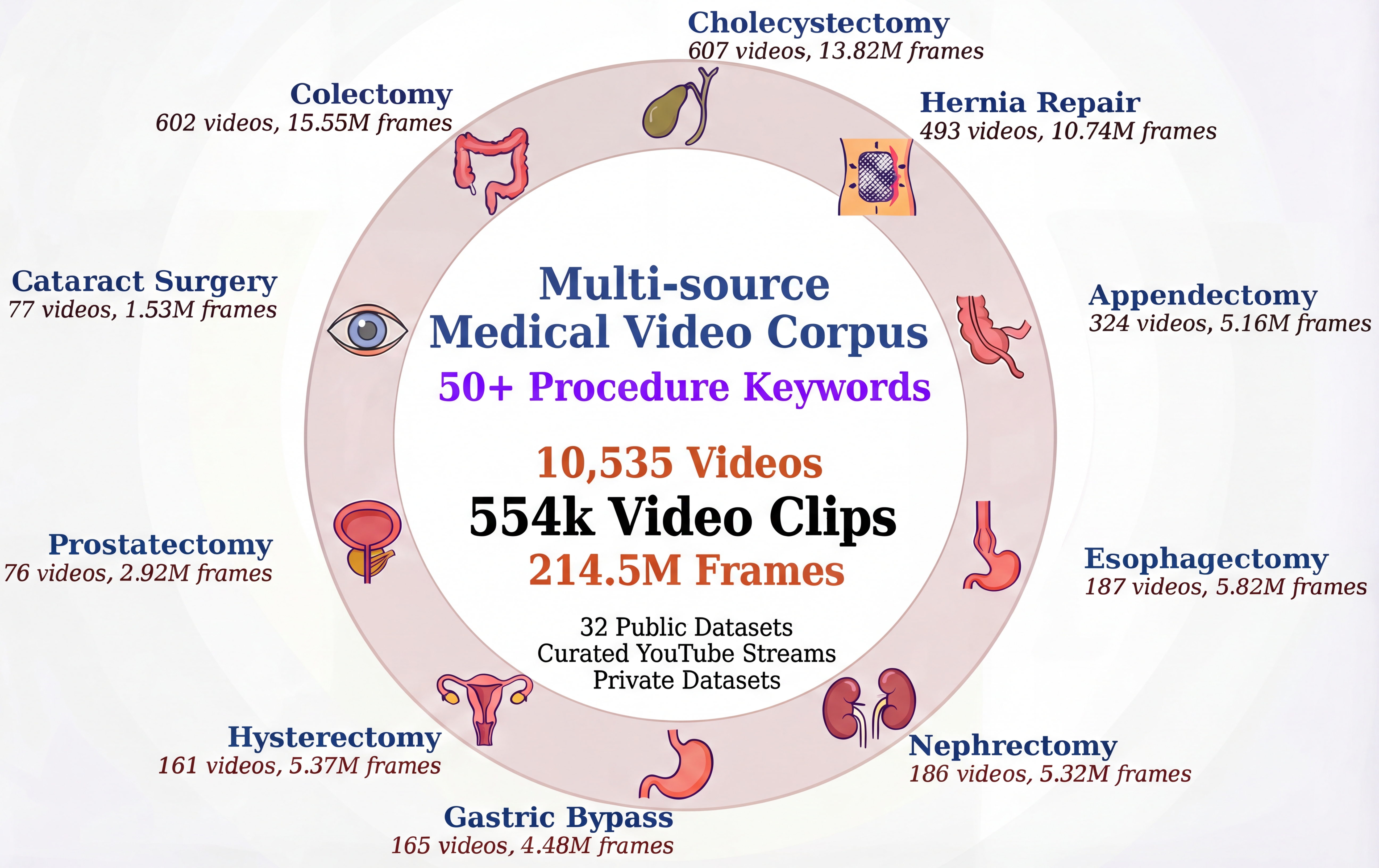}
        \caption{Distribution visualization of the pretraining corpus.}
        \label{fig:corpus_distribution}
    \end{minipage}
    \hfill
    \begin{minipage}[b]{0.54\textwidth}
        \centering
        \includegraphics[width=\textwidth]{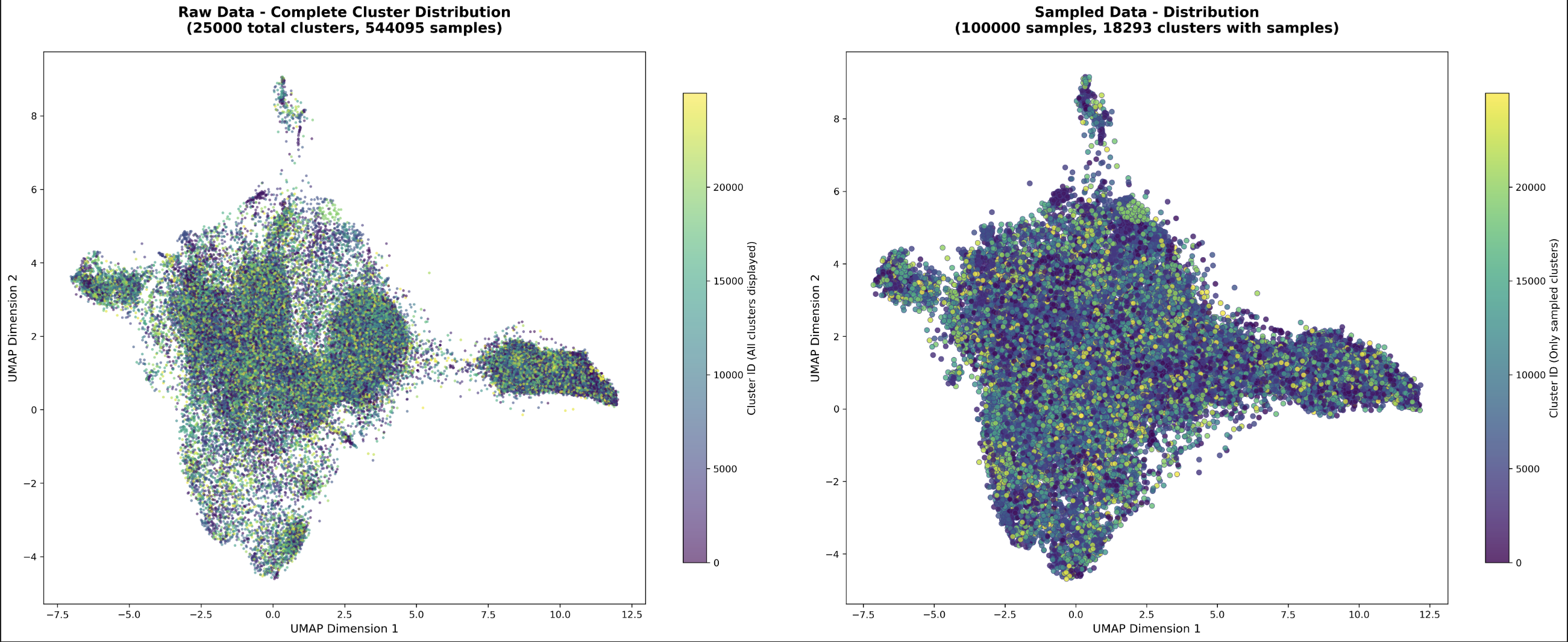}
        \caption{\textbf{Balanced data sampling.} UMAP projection of embeddings extracted via DINOv3 (left: raw, right: sampled).}
        \label{fig:umap_sampling}
    \end{minipage}
\end{figure}

The public component integrates 32 open datasets spanning four major clinical domains:
\textbf{Laparoscopy} (AIxSuture \cite{aixsuture}, AutoLaparo \cite{autolaparo}, Cholec80 / T45 / T50 \cite{endonet,cholect45,cholect50}, Endoscapes-CVS \cite{endoscapes}, HeiChole \cite{heichole}, M2CAI16 \cite{m2cai_lapchole}, MultiBypass140 \cite{multibypass140}, PmLR50 \cite{pmlr50}, SimSurgSkill2021 \cite{simsurgskill}, SurgicalActions160 \cite{surgicalactions160});
\textbf{Endoscopy} (Colonoscopic \cite{colonoscopic}, CVC-12k/ClinicDB~\cite{cvc12k,cvcclinicdb}, Endovis2019~\cite{endovis2019}, HyperKvasir \cite{hyperkvasir}, KUMC \cite{kumc}, Kvasir-Capsule \cite{kvasir_capsule}, LDPolypVideo \cite{ldpolypvideo}, PitVis \cite{pitvis}, PolypDiag~\cite{polypdiag}, PolypsSet \cite{polypsset});
\textbf{Cataract} (CATARACTS-1k \cite{cataracts}, Cataract-101/21~\cite{cataract101,cataract21});
\textbf{Robotic} (GraSP \cite{grasp}, JIGSAWS \cite{jigsaws}, PSI-AVA \cite{psiava}, SAR-RARP50 \cite{sarrarp50}); and \textbf{Mixed} (AVOS \cite{avos}).
The open-access educational stream contributes a long-tail taxonomy of more than 50 procedure keywords. After harmonization, the reported pretraining inventory contains 10,535 videos, 554K video clips, 
and 214.5M frames. This total consists of 2,790 videos (39.9M frames) from 
32 public datasets and 7,745 videos (174.6M frames) curated from publicly 
available platforms. We preserve cross-source heterogeneity to include realistic shifts in instrument scale, smoke, and illumination. To contextualize the scale, Table~\ref{tab:data_scale_comparison} compares our corpus against contemporaneous surgical foundation models. (See Fig.~\ref{fig:corpus_overview} for the full pipeline, Fig.~\ref{fig:corpus_distribution} for data distributions, and Sec.~\ref{sec:pretrain_impl} for details.)

\begin{table*}[t]
\caption{Pretraining data scale comparison across recent medical foundation models.}
\label{tab:data_scale_comparison}
\centering
{\footnotesize
\setlength{\tabcolsep}{6pt}
\renewcommand{\arraystretch}{1.05}
\begin{tabular*}{\textwidth}{@{\extracolsep{\fill}}l | l | l@{}}
\hline
\textbf{Method} & \textbf{Domain Focus} & \textbf{Reported Scale} \\
\hline
EndoDINO & GI Endoscopy & 100K -- 10M images \\
SurgeNetXL & General Surgery & ${>}$4.7M frames \\
SurgBench-P & Surgery (11 specialties) & 53M frames \\
EndoMamba & Endoscopy & 75K clips / 11.3M frames \\
SurgVLP & Surgical Lectures & 1,400 videos \\
SurgVISTA & Surgery ($>$20 procedures) & 3.6K videos / 3.55M frames \\
UniSurg & Universal Surgery & 3,658 hours \\
\hline
\textbf{Ours} & \textbf{Multi-Domain (4 major)} & \textbf{10.5K videos / 214.5M frames} \\
\hline
\end{tabular*}
}
\end{table*}

\subsection{Ethics Statement}
All public datasets are used in accordance with their respective licenses and terms of use. The open-access educational videos are publicly available, de-identified instructional materials containing no protected health information (PHI), and therefore require no IRB approval. Videos failing internal quality or privacy checks are excluded.

\section{Methods}
\subsection{Pretraining Implementation}\label{sec:pretrain_impl}
% We pretrain two video representation paradigms, SurgRec-MAE and SurgRec-JEPA, under a unified data curation, preprocessing, and training schedule. To construct a diversity-balanced unlabeled subset for pretraining, we subsample the unlabeled pool using a hierarchical clustering-based balanced sampling strategy following Vo \emph{et al.}~\cite{ssl_data_curation}. Specifically, we extract 768-d DINOv3 embeddings for $\sim$554k clips, perform hierarchical K-means clustering ($25\mathrm{k}\rightarrow5\mathrm{k}\rightarrow1\mathrm{k}$), and allocate a 10\% sampling budget \emph{top-down} to select samples nearest to each leaf centroid. For spatial data augmentation, all video frames are first resized such that the shortest side is 320 pixels while maintaining the original aspect ratio, followed by extracting random $224 \times 224$ spatial crops to construct the final video clips.
We pretrain SurgRec-MAE and SurgRec-JEPA under a unified data curation, preprocessing, and training schedule. For preprocessing, all frames are resized so that the shortest side is 320 pixels, followed by random $224{\times}224$ spatial crops. For data curation, we construct a balanced subset from the unlabeled pool following Vo~\emph{et al.}~\cite{ssl_data_curation}: we extract 768-d DINOv3 embeddings for $\sim$554k clips, apply hierarchical K-means clustering ($25\text{k}{\rightarrow}5\text{k}{\rightarrow}1\text{k}$), and select samples nearest to each leaf centroid under a 10\% top-down sampling budget.

During pretraining, models are optimized using a mixed-batch strategy combining a balanced unlabeled pool (primarily curated web videos) and a clinical core of 2,790 videos from the 32 public datasets. Batches are sampled with 15\% drawn exclusively from the clinical core and 85\% from a mixture of 70\% unlabeled and 30\% clinical data, yielding an effective global ratio of 59.5\% unlabeled and 40.5\% clinical. Pretraining utilizes 8 NVIDIA A100 GPUs ({\raise.17ex\hbox{$\scriptstyle\sim$}}8 days for SurgRec-MAE, 15 days for SurgRec-JEPA), while downstream fine-tuning takes 1--12 hours per dataset on 8 RTX 3090 GPUs.

% \subsection{Transfer Protocol}
% We evaluate transfer under two protocols. First, for visual representation transfer, pretrained self-supervised backbones are adapted to downstream datasets under the split standards described in Sec.~\ref{sec:benchmark_suite}. Second, for multimodal comparison, we perform a VLM protocol on the same test sets, where no task-specific fine-tuning is used for the VLM baselines.

\subsection{Evaluation Setup}
We evaluate under two settings. For SSL models, pretrained backbones are fine-tuned on each downstream dataset following the data splits described in Sec.~\ref{sec:benchmark_suite}. For VLM baselines, we perform zero-shot inference on the same test sets without any task-specific fine-tuning.

\section{Experiments}
% \subsection{Downstream Benchmark Suite}\label{sec:benchmark_suite}
% We evaluate transfer on a multi-procedure benchmark spanning four surgical domains. To ensure reproducibility and prevent temporal leakage, all partitions are strictly fixed at the video (case) level prior to clip extraction. Our deterministic split policy follows three regimes: (1) \emph{Official}: preserving benchmark-provided partitions (e.g., predefined folds in MultiBypass140); (2) \emph{Official+ours}: applying case-level partitioning after annotation-guided slicing for phase-oriented datasets lacking canonical clip-based splits; and (3) \emph{Ours}: a standard 7:2:1 video-level split where no community standard exists. All final split manifests are versioned and publicly released.
\subsection{Downstream Benchmark}\label{sec:benchmark_suite}
Our benchmark spans 16 datasets across four surgical domains. To ensure reproducibility, train and test splits are defined at the video (case) level so that no two splits share clips from the same video. Split assignment follows a three-tier priority rule: (1) \emph{Official}: official benchmark partitions are used when available (e.g., predefined folds in MultiBypass140); (2) \emph{Community}: publicly released community splits are adopted when no official partition exists; and (3) \emph{Ours}: a standard 7:2:1 video-level split is applied where neither official nor community standards exist. All split manifests are versioned and publicly released.

\subsection{Baselines}
We benchmark SurgRec-MAE and SurgRec-JEPA against two families of baselines. \textbf{SSL baselines} include the general-domain models VideoMAE~\cite{videomae}, JEPA~\cite{vjepa2}, and DINOv3~\cite{dinov3}, as well as DINOv3-SurgeNetXL, a domain-adapted reference obtained by continuing pretraining DINOv3 on our clinical subset; full results are listed in Table~\ref{tab:main_comparison_ssl}. \textbf{VLM baselines} include Qwen3-VL-8B, LLaVA-NeXT-7b, and Qwen2.5-VL-7B, evaluated on the same test sets.

\section{Results}

\begin{table}[t]
\caption{Main comparison on Acc@1 (\%) against SSL baselines. SR and V-MAE denote SurgRec and VideoMAE, respectively. \texttt{SR-MAE (w/o bal.)} refers to SurgRec-MAE trained without balanced sampling. Dataset names are abbreviated (e.g., Cat: Cataract, Colono: Colonoscopic).}
\label{tab:main_comparison_ssl}
\centering
{\footnotesize 
\setlength{\tabcolsep}{2pt} 
\renewcommand{\arraystretch}{1.1}
\begin{tabularx}{\linewidth}{@{} l | *{5}{>{\centering\arraybackslash}X} | *{2}{>{\centering\arraybackslash}X} @{}}
\hline
\multirow{2}{*}{\textbf{Dataset}} & \multicolumn{5}{c|}{\textbf{Baselines}} & \multicolumn{2}{c}{\textbf{Ours}} \\
\cline{2-8}\noalign{\vskip 1pt}
& \multicolumn{1}{c}{\shortstack[c]{\textbf{DINOv3}\\\phantom{\textbf{Surg}}}}
& \multicolumn{1}{c}{\shortstack[c]{\textbf{DINOv3}\\\textbf{Surg}}}
& \multicolumn{1}{c}{\shortstack[c]{\textbf{V-MAE}\\\phantom{\textbf{Surg}}}}
& \multicolumn{1}{c}{\shortstack[c]{\textbf{SR-MAE}\\\textbf{(w/o bal.)}}}
& \multicolumn{1}{c}{\shortstack[c]{\textbf{JEPA}\\\phantom{\textbf{Surg}}}}
& \multicolumn{1}{c}{\shortstack[c]{\textbf{SR}\\\textbf{MAE}}}
& \multicolumn{1}{c}{\shortstack[c]{\textbf{SR}\\\textbf{JEPA}}} \\
\hline
AIxSuture  & 32.61 & 39.13 & 39.13 & 39.13 & 39.13 & \textbf{43.48} & 39.13 \\
AutoLap.   & 17.74 & 19.35 & 19.35 & 17.74 & 20.97 & \textbf{22.58} & 17.74 \\
Cat-21     & 15.69 & 15.69 & 15.69 & 15.69 & 15.69 & 15.69 & 15.69 \\
Cat-101    & 17.97 & 19.53 & 19.53 & 21.09 & 24.22 & \textbf{35.16} & 23.44 \\
CAT-1k     & 9.90  & 14.85 & 11.88 & 16.83 & 19.80 & \textbf{18.81} & 17.82 \\
Cholec80   & 22.34 & 16.12 & 14.65 & 24.18 & 26.01 & 29.30 & \textbf{31.14} \\
Colono.    & 53.33 & 53.33 & 53.33 & 53.33 & 53.33 & \textbf{73.33} & 53.33 \\
HyperK.    & 43.03 & 41.81 & 25.11 & 36.91 & 41.65 & \textbf{55.28} & 42.27 \\
JIGSAWS    & 18.12 & 17.11 & 15.67 & 21.23 & 29.72 & \textbf{47.49} & 32.24 \\
Kvasir-C.  & 46.34 & 47.56 & 19.51 & 46.34 & 48.78 & \textbf{73.17} & 60.98 \\
LapGyn4    & 62.68 & 65.63 & 62.68 & 67.07 & 60.66 & \textbf{68.13} & 64.13 \\
LDPoly.    & 90.35 & 90.35 & 90.35 & 90.35 & 90.35 & \textbf{90.76} & 90.42 \\
M2CAI16    & 23.01 & 25.66 & 12.39 & 30.09 & 26.55 & \textbf{38.05} & \textbf{38.05} \\
MultiByp.  & 38.62 & 42.63 & 20.54 & 51.56 & 12.88 & \textbf{56.92} & 56.47 \\
SurgAct.   & 6.25  & 6.25  & 6.25  & \textbf{18.75} & 12.50 & \textbf{18.75} & 6.25 \\
SAR-RARP   & 23.01 & 26.88 & 25.97 & 24.37 & 22.10 & \textbf{37.59} & 36.67 \\
\hline
\end{tabularx}
}
\end{table}

\begin{table}[t]
\caption{VLM Acc@1 (\%). \textbf{SurgRec-MAE} is the supervised reference. P1: baseline prompt; P2: stricter output constraint. $\Delta$ = P2$-$P1 (points). Dataset names are abbreviated as in Table \ref{tab:main_comparison_ssl}.}
\label{tab:vlm_comprehensive}
\centering
{\footnotesize
\setlength{\tabcolsep}{2.5pt} 
\renewcommand{\arraystretch}{1.1}
\begin{tabular}{@{} l | c | c c c | c c c | c c c @{}}
\hline
\multirow{2}{*}{\textbf{Dataset}} &
\multicolumn{1}{c|}{\textbf{Reference}} &
\multicolumn{3}{c|}{\makecell{\textbf{Qwen3-}\\\textbf{VL-8B}}} &
\multicolumn{3}{c|}{\makecell{\textbf{LLaVA-}\\\textbf{NeXT-7B}}} &
\multicolumn{3}{c}{\makecell{\textbf{Qwen2.5-}\\\textbf{VL-7B}}} \\
\cline{2-11}
& \makecell{\textbf{SurgRec}\\\textbf{-MAE}} &
\textbf{P1} & \textbf{P2} & $\boldsymbol{\Delta}$ &
\textbf{P1} & \textbf{P2} & $\boldsymbol{\Delta}$ &
\textbf{P1} & \textbf{P2} & $\boldsymbol{\Delta}$ \\
\hline
AIxSuture & \textbf{43.48} & 28.26 & 36.96 & +8.70 & 21.74 & 19.57 & -2.17 & 36.96 & 39.13 & +2.17 \\
AutoLap.  & \textbf{22.58} & 19.35 & 22.58 & +3.23 & 11.29 & 11.29 & +0.00 & 12.90 & 14.52 & +1.61 \\
Cat-21    & \textbf{15.69} & 7.84  & 3.92  & -3.92 & 9.80  & 11.76 & +1.96 & 13.73 & 13.73 & +0.00 \\
Cat-101   & \textbf{35.16} & 13.28 & 11.72 & -1.56 & 8.59  & 8.59  & +0.00 & 10.94 & 11.72 & +0.78 \\
CAT-1k    & \textbf{18.81} & 4.95  & 4.95  & +0.00 & 5.94  & 3.96  & -1.98 & 5.94  & 5.94  & +0.00 \\
Cholec80  & \textbf{29.30} & 23.08 & 25.64 & +2.56 & 14.65 & 15.02 & +0.37 & 19.05 & 20.88 & +1.83 \\
Colono.   & \textbf{73.33} & 40.00 & 40.00 & +0.00 & 53.33 & 53.33 & +0.00 & 40.00 & 40.00 & +0.00 \\
HyperK.   & \textbf{55.28} & 16.06 & 18.04 & +1.99 & 23.39 & 24.46 & +1.07 & 8.87  & 10.86 & +1.99 \\
JIGSAWS   & \textbf{47.49} & 13.76 & 14.18 & +0.42 & 8.48  & 9.79  & +1.30 & 14.47 & 14.83 & +0.36 \\
Kvasir-C. & \textbf{73.17} & 8.54  & 10.98 & +2.44 & 19.51 & 14.63 & -4.88 & 20.73 & 21.95 & +1.22 \\
LapGyn4   & \textbf{68.13} & 19.00 & 17.84 & -1.16 & 8.29  & 5.79  & -2.51 & 14.13 & 14.22 & +0.10 \\
LDPoly.   & \textbf{90.76} & 85.62 & 85.49 & -0.14 & 9.65  & 9.65  & +0.00 & 88.40 & 89.03 & +0.63 \\
M2CAI16   & \textbf{38.05} & 33.63 & 35.40 & +1.77 & 16.81 & 12.39 & -4.42 & 19.47 & 21.24 & +1.77 \\
MultiByp. & \textbf{56.92} & 2.46  & 2.90  & +0.45 & 1.79  & 1.79  & +0.00 & 6.25  & 6.92  & +0.67 \\
SAR-RARP  & \textbf{37.59} & 28.93 & 29.84 & +0.91 & 28.25 & 26.88 & -1.37 & 25.74 & 25.97 & +0.23 \\
SurgAct.  & \textbf{18.75} & 6.25  & 0.00  & -6.25 & 12.50 & 12.50 & +0.00 & 6.25  & 6.25  & +0.00 \\
\hline
\end{tabular}
}
\end{table}

% \subsection{Overall Transfer Performance}
% To validate the benefit of large-scale in-domain pretraining, we first evaluate the absolute performance gains across our 16-dataset benchmark. Table~\ref{tab:main_comparison_ssl} reports the full benchmark values across all 16 datasets. To better understand the paired improvements between our domain-pretrained checkpoints (\texttt{SurgRec-MAE} and \texttt{SurgRec-JEPA}) and their original generic counterparts (VideoMAE and JEPA), we aggregate these results into domain-level macro scores in Table~\ref{tab:domain_macro_combined}. Overall, in-domain pretraining yields positive average gains across the benchmark (\(+3.18\) points for VideoMAE and \(+2.61\) points for JEPA), although absolute improvements remain dataset-dependent due to varying clinical domain shifts. By analyzing both the full matrix and the domain-level aggregations, the transfer benefits are visible under a unified evaluation protocol.

\subsection{Main Results}
We first evaluate SurgRec-MAE and SurgRec-JEPA across the full 16-dataset benchmark to assess how well surgical-domain pretraining scales to diverse downstream tasks. Table~\ref{tab:main_comparison_ssl} reports per-dataset accuracy for all models under a unified evaluation. We further aggregate these results into domain-level macro scores in Table~\ref{tab:domain_macro_combined}, where we directly compare SurgRec-MAE and SurgRec-JEPA against their general-domain counterparts VideoMAE and JEPA.
Overall, SurgRec variants yield consistent improvements across the benchmark (\(+3.18\) points for SurgRec-MAE over VideoMAE and \(+2.61\) points for SurgRec-JEPA over JEPA), indicating that surgical-domain pretraining leads to representations that generalize well across heterogeneous procedures and clinical settings. While per-dataset gains vary in magnitude, the improvements hold consistently across both model families and all four clinical domains, suggesting that the benefits are robust and not limited to any particular procedure or dataset.

\subsection{VLM Comparison}
We compare SurgRec against state-of-the-art VLMs to demonstrate the advantages of surgical video pretraining. SurgRec-MAE consistently and substantially outperforms all VLM baselines in fine-grained surgical recognition across all 16 datasets (Table~\ref{tab:vlm_comprehensive}), as our pretrained representations capture procedure-specific temporal patterns that generic instruction tuning fails to model. Beyond raw performance, SurgRec exhibits stable behavior across datasets, whereas VLMs suffer from notable prompt sensitivity: minor phrasing modifications (e.g., emphasizing an \textit{expert} role in P2) cause noticeable performance fluctuations ($\Delta$) across datasets. This instability further highlights that while VLMs offer broad open-vocabulary capabilities, pretraining on surgical video corpora yields more robust and reliable representations for downstream surgical tasks.

\subsection{Generalization Analysis}
We evaluate generalization using a macro-averaged protocol to reduce sensitivity to domain imbalance (e.g., the dominance of laparoscopy data). As shown in Table~\ref{tab:domain_macro_combined}, SurgRec-MAE achieves a +5.43 point gain in \textit{Overall Macro} Acc@1 over the VideoMAE baseline, while simultaneously improving the \textbf{Worst Domain} score (Cataract) from 19.09\% to 22.23\%, indicating that performance gains are not concentrated in any single domain. SurgRec-JEPA similarly yields a +1.91 point macro improvement over its generic counterpart, primarily driven by gains in Endoscopy and Laparoscopy. Together, these results show that scaling diverse surgical video pretraining with balanced sampling improves both average performance and worst-case domain coverage simultaneously, neither of which is achieved by training on imbalanced data or relying on generic pretraining alone.

\begin{table}[t]
\caption{Generalization analysis on the 4-domain benchmark (Acc@1, \%). Domain scores are macro-averaged; \textit{Overall Macro} and \textit{Worst Domain} denote the mean and minimum across domains, respectively.}
\label{tab:domain_macro_combined}
\centering
{\footnotesize
\setlength{\tabcolsep}{1pt} 
\renewcommand{\arraystretch}{1.1}
\begin{tabular}{@{} l|c|c|c|c|c|c @{}}
\hline
\textbf{Model} & \textbf{Cataract} & \textbf{Robotic} & \makecell{\textbf{Endoscopy}} & \makecell{\textbf{Laparoscopy}} & \makecell{\textbf{Overall}\\\textbf{Macro}} & \makecell{\textbf{Worst}\\\textbf{Domain}} \\
\hline
\multicolumn{7}{@{}l}{\textbf{VideoMAE Variants}} \\
\hline
V-MAE (Baseline) & 19.09 & 42.22 & 62.12 & 29.05 & 38.12 & 19.09 \\
SR-MAE (w/o bal.) & 17.87 & 22.80 & 56.73 & 35.50 & 33.23 & 17.87 \\
SR-MAE (Ours) & \textbf{22.23} & \textbf{47.37} & \textbf{71.49} & \textbf{33.11} & \textbf{43.55} & \textbf{22.23} \\
\textit{$\Delta$ vs. Base} & \textit{+3.14} & \textit{+5.15} & \textit{+9.37} & \textit{+4.06} & \textit{+5.43} & \textit{+3.14} \\
\textit{$\Delta$ vs. w/o bal.} & \textit{+4.36} & \textit{+24.57} & \textit{+14.76} & \textit{-2.39} & \textit{+10.32} & \textit{+4.36} \\
\hline
\multicolumn{7}{@{}l}{\textbf{JEPA Variants}} \\
\hline
JEPA (Baseline) & 19.50 & 31.52 & 57.39 & 25.50 & 33.48 & 19.50 \\
SR-JEPA (Ours) & 18.84 & 30.98 & \textbf{61.10} & \textbf{30.62} & \textbf{35.39} & 18.84 \\
\textit{$\Delta$ vs. Base} & \textit{-0.66} & \textit{-0.54} & \textit{+3.71} & \textit{+5.12} & \textit{+1.91} & \textit{-0.66} \\
\hline
\end{tabular}
}
\end{table}

\subsection{Effect of Balanced Sampling}
To isolate the effect of data distribution from raw data scale, we compare \textbf{SurgRec\allowbreak-MAE\allowbreak~w/o\allowbreak~bal.} against a version trained on the raw, unbalanced corpus (\texttt{SurgRec-MAE w/o bal.}) under identical computational budgets. Without balancing, the training distribution is heavily skewed toward laparoscopy and GI endoscopy. As shown in Table~\ref{tab:domain_macro_combined}, balanced sampling yields substantial gains in the \textbf{Robotic} (+24.57 points) and \textbf{Endoscopy} (+14.76 points) domains, and improves the \textit{Worst Domain} score by +4.36 points. These results confirm that generalization gains stem not only from increased data volume, but critically from our balanced sampling strategy.

% \section{Discussion}
% The core finding of this study is that a data-centric scaling strategy, empowered by balanced sampling, is highly effective for medical video foundation modeling. Our ablation reveals that mitigating long-tail procedural bias is as crucial as increasing raw data volume, significantly improving robustness to domain shifts. Furthermore, our benchmark highlights the fundamental limitations of general-purpose Vision-Language Models (VLMs) in specialized surgical video analysis. While VLMs possess broad open-vocabulary capabilities, they suffer from prompt instability and substantially underperform on fine-grained temporal reasoning tasks. In contrast, self-supervised models pretrained on domain-specific corpora consistently capture procedure-specific dynamics, demonstrating that dedicated in-domain pretraining remains essential for reliable surgical understanding.

% This study has several limitations. Leveraging web-based videos introduces a potential domain gap, as they often under-represent adverse events and unedited idle periods. Additionally, some benchmark subsets remain small or noisy. Future work will target multi-center evaluations in strictly low-data, unedited clinical settings.

\section{Discussion}
Our results show that two factors jointly drive strong surgical video representations: data scale and balanced sampling. While scaling the corpus is important, our ablation confirms that mitigating procedural imbalance is equally critical, as simply increasing data volume without balancing yields substantially weaker cross-domain generalization. Our benchmark further reveals a clear gap between domain-pretrained encoders and general-purpose VLMs: VLMs suffer from prompt instability and consistently underperform on fine-grained temporal recognition, whereas pretraining on surgical video corpora produces representations that transfer reliably across heterogeneous procedures and clinical settings.

This study has several limitations. Web-sourced videos introduce a domain gap, as they typically differ from real clinical recordings in content coverage, often omitting intraoperative complications and consisting predominantly of highlight footage. Additionally, some benchmark subsets remain small or noisy, which may affect the reliability of performance. Future work will target evaluations in more realistic, multi-center clinical settings with diverse patient populations.

\section{Conclusion}
% This work demonstrates that scaling a diverse, multi-source video corpus is a fundamental driver for robust medical video foundation models. We show that explicitly balancing procedural heterogeneity yields massive gains in cross-domain generalization. Furthermore, our benchmark proves that domain-pretrained self-supervised encoders are vastly superior and more reliable than VLMs for fine-grained surgical temporal understanding. Ultimately, balanced pretraining is an indispensable step toward deploying reusable AI assets capable of withstanding complex real-world clinical distribution shifts.

This work shows that scaling diverse multi-source video data is key to building robust medical video foundation models. Explicitly balancing procedural heterogeneity significantly improves cross-domain generalization. Our benchmark also demonstrates that domain-pretrained self-supervised encoders outperform VLMs for fine-grained surgical temporal understanding. Overall, balanced pretraining is essential for developing reliable, reusable AI systems that adapt to real-world clinical distribution shifts.

%% Removed for anonymized MICCAI submission:
%% Acknowledgments and disclosure can be added back for camera-ready.

\end{document}